\def\year{2019}\relax
\documentclass[letterpaper]{article} %DO NOT CHANGE THIS
\usepackage{aaai19}  %Required
\usepackage{times}  %Required
\usepackage{helvet}  %Required
\usepackage{courier}  %Required
\usepackage{url}  %Required
\usepackage{graphicx}  %Required

\usepackage{latexsym}
\usepackage{amsmath}
\usepackage{amssymb}
\usepackage{xcolor}
\usepackage{enumerate}
\usepackage{comment}
\usepackage{subfigure}
\usepackage{multirow}
\usepackage{bm}
\usepackage{url}

\begin{document}
\title{$\rm Y^2$Seq2Seq: Cross-Modal Representation Learning for 3D Shape and Text by Joint Reconstruction and Prediction of View and Word Sequences}
\author{Zhizhong Han\textsuperscript{1,2},
Mingyang Shang\textsuperscript{1},
Xiyang Wang\textsuperscript{1},
Yu-Shen Liu\textsuperscript{1}\thanks{Corresponding author: Yu-Shen Liu},
Matthias Zwicker\textsuperscript{2}\\
\textsuperscript{1}{School of Software, Tsinghua University, Beijing, China}\\
\textsuperscript{2}{Department of Computer Science, University of Maryland, College Park, USA}\\
h312h@umd.edu,
smy16@mails.tsinghua.edu.cn,
wangxiya16@mails.tsinghua.edu.cn,
liuyushen@tsinghua.edu.cn,
zwicker@cs.umd.edu
}
%\author{AAAI Press\\
%Association for the Advancement of Artificial Intelligence\\
%2275 East Bayshore Road, Suite 160\\
%Palo Alto, California 94303\\
%}
\maketitle
\begin{abstract}
\begin{quote}
Jointly learning representations of 3D shapes and text is crucial to support tasks such as cross-modal retrieval or shape captioning.
%explore 3D shapes with specific properties among large scale dataset using language.
A recent method employs 3D voxels to represent 3D shapes, but this limits the approach to low resolutions due to the computational cost caused by the cubic complexity of 3D voxels. Hence the method suffers from a lack of detailed geometry. To resolve this issue, we propose $\rm Y^2Seq2Seq$, a view-based model, to learn cross-modal representations by joint reconstruction and prediction of view and word sequences. Specifically, the network architecture of $\rm Y^2Seq2Seq$ bridges the semantic meaning embedded in the two modalities by two coupled ``Y'' like sequence-to-sequence (Seq2Seq) structures. In addition, our novel hierarchical constraints further increase the discriminability of the cross-modal representations by employing more detailed discriminative information. Experimental results on cross-modal retrieval and 3D shape captioning show that $\rm Y^2Seq2Seq$ outperforms the state-of-the-art methods.
\end{quote}
\end{abstract}

\section{Introduction}
With the development of 3D modeling and scanning techniques, more and more 3D shapes become available on the Internet with detailed physical properties, such as texture, color, and material. With large 3D datasets, however, shape class labels are becoming too coarse of a tool to help people efficiently find what they want, and visually browsing through shape classes is cumbersome. To alleviate this issue, an intuitive approach is to allow users to describe the desired 3D object using a text description. Jointly understanding 3D shape and text by learning a cross-modal representation, however, is still a challenge because it requires an efficient 3D shape representation that can capture highly detailed 3D shape structures.

To overcome this challenge, a 3D-Text cross-modal dataset was recently released in~\cite{chenkevin2018}, where a combined multimodal association model was also proposed to capture the many-to-many relations between 3D voxels and text descriptions. This model employed a shape encoder to compute the embeddings of 3D shapes directly from 3D voxels. However, this strategy is limited to learning from low resolution voxel representations due to the computational cost caused by the cubic complexity of 3D voxels. This leads to low discriminability of learned cross-modal representations due to the lack of detailed geometry information.

We resolve this issue by proposing to learn cross-modal representations of 3D shape and text from view sequences and word sequences, where each 3D shape is represented by a view sequence. Our deep learning model captures the correlation between 3D shape and text by the joint reconstruction and prediction of view and word sequences from each modality. This strategy aims to get part-level information as much as possible to understand mainly part-related descriptions, which remedies the lack of part prior knowledge in 3D-Text understanding. We call our model $\rm Y^2$Seq2Seq, since it is implemented by two coupled ``Y'' like sequence-to-sequence (Seq2Seq) structures. In addition, we employ novel hierarchical constraints that can be easily added to extract more detailed discriminative information from different aspects. Specifically, $\rm Y^2$Seq2Seq consists of a 3D shape branch and a text branch. Starting from the encoding of either 3D shape modality or text modality, each branch jointly reconstructs the modality itself and predicts the counterpart modality using one RNN encoder and two RNN decoders. The two branches are coupled by sharing parameters in decoders for reconstruction and prediction, which is effective to bridge the semantic meaning embedded in 3D shape modality and text modality. In addition, we propose to employ the discriminative information at the class level, instance pair level, and instance level as constraints in the training procedure. Our significant contributions are list below.

\begin{itemize}
\item We propose a deep learning model called $\rm Y^2$Seq2Seq, which enables to learn cross-modal representations of 3D shape and text from view sequences and word sequences.
\item Our novel coupled ``Y'' like Seq2Seq structures have a powerful capability to bridge the semantic meaning of two sequence-represented modalities by joint reconstruction and prediction.
\item Our results demonstrate that our novel hierarchical constraints can further increase the discriminability of learned cross-modal representations by employing more detailed discriminative information.
\end{itemize}

\section{Related work}
\noindent\textbf{Joint 3D-Text representation learning. }In a recent pioneering study, a combined multimodal association model~\cite{chenkevin2018} was proposed to jointly understand 3D shape and text based on a novel 3D-Text cross-modal dataset. This model employs a CNN+RNN and a 3D-CNN to extract single-modal features of text and 3D shape respectively, where the 3D-CNN learns from 3D voxels. Then, similarities within each modality and cross modality are learned by a metric learning method. However, due to the computational cost caused by the cubic complexity of 3D voxels, this model is limited to learning from low resolution voxels, and the lack of detailed geometry information affects the discriminability of the learned joint representations. To resolve this issue, $\rm Y^2Seq2Seq$ aims to learn features of 3D shapes from view sequences.

\noindent\textbf{Joint 2D sequence-Text representation learning. }Learning from view sequences makes $\rm Y^2Seq2Seq$ related to joint 2D sequence-Text representation learning, such as video captioning~\cite{xushen2018aaai,BairuiCVPR2018,Venugopalan2015} and the recent GIF-Text cross-modal retrieval~\cite{YaleSongGIF2018}.

Besides the rendered views that we use rather than natural images, the main difference between $\rm Y^2Seq2Seq$ and these studies is that $\rm Y^2Seq2Seq$ is required to focus more on part-level understanding rather than object-level understanding. This is because our approach works on individual 3D objects, and there is only one object in a view. Hence, the corresponding representation mainly captures information about parts of this object.
%MZ: the following sentence is hard to understand. I also think the argument is not very convincing. How could you show experimentally that this is actually true?
%This makes rendered view based text understanding is more difficult than natural image based text understanding, since the former can get no part prior information from DNN, such as VGG, pretrained under large scale object-level dataset, such as ImageNet, as the object prior information for the latter.
For better joint understanding of rendered view and text, $\rm Y^2Seq2Seq$ employs two coupled ``\rm Y'' like Seq2Seq branches, and each branch simultaneously conducts joint reconstruction and prediction in both modalities. This novel structure is able to learn from discriminative information in both modalities as much as possible, which we also facilitate by including additional constraints. In addition, another difference between $\rm Y^2Seq2Seq$ and these studies is that we aim to explicitly learn cross-modal representations, which enables $\rm Y^2Seq2Seq$ for cross-modal retrieval and captioning at the same time, rather than merely retrieval~\cite{YaleSongGIF2018} or captioning~\cite{BairuiCVPR2018,Venugopalan2015}.

In terms of the model structure, $\rm Y^2Seq2Seq$ is similar to the prototype bimodal deep autoencoder~\cite{NgiamICML2011} which was also adopted by Shared Latent Representation (SLR) learning with variational autoencoder background~\cite{xushen2018aaai}. However, these models contain only a single branch, which makes it hard to employ additional constraints on the learned cross-modal representations. The encoder-decoder-reconstructor structure~\cite{BairuiCVPR2018} also suffers from this issue. The correspondence fully-modal autoencoder~\cite{FangxiangACMM2014} also employed a similar structure to $\rm Y^2Seq2Seq$ for image captioning. However, this method cannot handle 2D sequences, and moreover, regards the sentence as a whole, which means that it could not learn the relationship between words and shape characteristics.

\section{$\rm Y^2Seq2Seq$}
\begin{figure}[!htb]
  \centering
  % the following command controls the width of the embedded PS file
  % (relative to the width of the current column)
  %\includegraphics[width=.95\linewidth, bb=39 696 126 756]{figures/definition3.eps}
   \includegraphics[width=\linewidth]{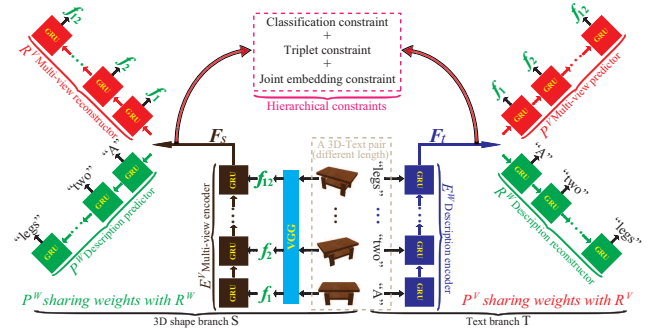}
  % replacing the above command with the one below will explicitly set
  % the bounding box of the PS figure to the rectangle (xl,yl),(xh,yh).
  % It will also prevent LaTeX from reading the PS file to determine
  % the bounding box (i.e., it will speed up the compilation process)
  % \includegraphics[width=.95\linewidth, bb=39 696 126 756]{sampleFig}
  %
  %
\caption{\label{fig:Framework} The framework of $\rm Y^2Seq2Seq$ consists of a 3D shape branch $\rm S$ and text branch $\rm T$. Each branch jointly reconstructs within the modality and predicts across modalities.}
\end{figure}

\noindent\textbf{Overview. }The framework of $\rm Y^2Seq2Seq$ is illustrated in Fig.~\ref{fig:Framework}. $\rm Y^2Seq2Seq$ is formed by a 3D shape branch $\rm S$ and a text branch $\rm T$. Each branch is a ``$\rm Y$'' like seq2seq model which is formed by one RNN encoder and two RNN decoders, and the two RNN decoders jointly reconstruct within the modality and predict across modalities. $\rm S$ and $\rm T$ are coupled by sharing weights involved in the decoders. Note that in Fig.~\ref{fig:Framework}, the ``$\rm Y$'' structure of the shape branch is rotated by 90 degrees in counter-clockwise orientation, and the one of the text branch is rotated in clockwise orientation.

For each pair of 3D shape $s$ and its description $t$, $(s,t)$ is learned in both branches. In the shape branch $\rm S$, the multi-view encoder $E^V$ aims to learn the feature $\bm{F}_s$ of $s$ by aggregating $N$ views $\bm{v}=[v_1,...,v_i,...,v_N]$, where $i\in[1,N]$. The first decoder in the shape branch is a multi-view reconstructor $R^V$ to reconstruct the low-level feature $\bm{f}_i$ of each view $v_i$. The second decoder is a description predictor $P^W$ to generate $t$ by predicting $M$  words $\bm{w}=[w_1,...,w_j,...,w_M]$, where $j\in[1,M]$. On the other hand, in the text branch $\rm T$, the description encoder $E^W$ aims to learn the feature $\bm{F}_t$ of $t$ by aggregating its $M$ words $\bm{w}$. The first decoder is a description reconstructor $R^W$ to reconstruct $\bm{w}$, and the second one is a multi-view predictor $P^V$ to predict the low-level feature $\bm{f}_i$ of each view $v_i$.

Finally, we include additional hierarchical constraints to learn the cross-modal representations $\bm{F}_s$ and $\bm{F}_t$. The constraints aim to capture the hierarchically discriminative information at the class level (using a classification constraint), at the instance pair level (using a triplet constraint), and at the instance level (using a joint embedding constraint).

\noindent\textbf{View sequence capturing. }A view sequence $\bm{v}$ is formed by uniformly capturing $N$ sequential views $v_i$ around a shape $m$ on a circle. We render shapes that are represented using 3D voxel grids. The cameras are elevated $30^{\circ}$ from the ground plane, pointing to the centroid of $m$. In our experiments, we set $N$ to 12.

\noindent\textbf{3D shape branch $\rm S$. }$\rm S$ is formed by a multi-view encoder $E^V$, a multi-view reconstructor $R^V$, and a description predictor $P^W$.

$E^V$ is implemented by a VGG19~\cite{Simonyan14c} and a RNN. The VGG19 first extracts low-level feature $\bm{f}_i$ of each view $v_i$ in the view sequence $\bm{v}$ from the last 4096 dimensional fully connected layer. Then, $E^V$ learns the feature $\bm{F}_s$ of shape $s$ by aggregating all the $N$ view features, where $\bm{F}_s$ is the hidden state at the $N$-th step of the RNN.

%with GRU cell~\cite{Cho2014On}

Starting from $\bm{F}_s$, $R^V$ and $P^W$, which are also implemented by a RNN, jointly reconstruct the view features in sequence $\bm{v}$ and predict the word sequence $\bm{w}$. The reconstructed view features $\bm{f}_i'$ and predicted word sequence $\bm{w}'$ are expected to be close to the ground truth view features $\bm{f}_i$ and word sequence $\bm{w}$. Thus, the loss of branch $\rm S$ is formed by the loss $L_{V2V}$ from view to view and the loss $L_{V2W}$ from view to word, as defined below,

\begin{equation}
\begin{aligned}
& L_{\rm S}=\alpha L_{V2V}+\beta L_{V2W}, \\
& L_{V2V}=\frac{1}{N}\sum_{i\in[1,N]}\|\bm{f}_i'-\bm{f}_i\|_2^2, \\
& L_{V2W}=-\sum_{j\in[1,M]}\log p(w_j|w_{<j},\bm{v}), \\
\end{aligned}
\end{equation}

\noindent where $w_j$ is the $j$-th word in the word sequence $\bm{w}$, $w_{<j}$ represents the words in front of $w_j$, $p(w_j|w_{<j},\bm{v})$ is the probability of correctly predicting $j$-th word according to the previous words $w_{<j}$ and the view sequence $\bm{v}$, and $\alpha$ and $\beta$ are two balance weights.

%\begin{equation}
%\label{eq:Sencoder}
%L_{\rm S}=L_{V2V}+L_{V2S}
%\end{equation}
%
%\begin{equation}
%\label{eq:Sdecoder1}
%L_{V2V}=\frac{1}{N}\sum_{i\in[1,N]}\|\bm{f}_i-\bm{f}_i'\|_2^2
%\end{equation}
%
%\begin{equation}
%\label{eq:Sdecoder2}
%L_{V2S}=-\sum_{j\in[1,M]}\log P(w_j'|w_{<j}',\bm{v})
%\end{equation}

\noindent\textbf{Text branch $\rm T$. }$\rm T$ is formed by a description encoder $E^W$, a description reconstructor $R^W$ and a multi-view predictor $P^V$. Note that $\rm T$ is coupled with $\rm S$ by sharing weights in $R^W$ with description predictor $P^W$ in $\rm S$ and sharing weights in $P^V$ with multi-view reconstructor $R^V$ in $\rm S$.

$E^W$, $R^W$ and $P^V$ are all implemented by a RNN. $E^W$ learns the feature $\bm{F}_t$ of the description $t$ of shape $s$ by aggregating the embedding $\bm{e}_j$ of words $w_j$ in sequence $\bm{w}$. The word embedding $\bm{e}_j$ is learned with the other parameters in $\rm Y^2Seq2Seq$ together. The feature $\bm{F}_t$ is represented by the hidden state at the $M$-th step of $E^W$.

Starting from $\bm{F}_t$, $R^W$ and $P^V$ jointly reconstruct the word sequence $\bm{w}$ and predict the view features in sequence $\bm{v}$. The reconstructed word sequence $\bm{w}''$ and predicted view features $\bm{f}_i''$ are expected to be close to the ground truth word sequence $\bm{w}$ and view features $\bm{f}_i$. Thus, the loss of branch $\rm T$ is formed by the loss $L_{W2W}$ from word to word and the loss $L_{W2V}$ from word to view, as defined below,

\begin{equation}
\begin{aligned}
& L_{\rm T}=\gamma L_{W2W}+\delta L_{W2V}, \\
& L_{W2W}=-\sum_{j\in[1,M]}\log p(w_j|w_{<j},\bm{w}), \\
& L_{W2V}=\frac{1}{N}\sum_{i\in[1,N]}\|\bm{f}_i''-\bm{f}_i\|_2^2, \\
\end{aligned}
\end{equation}

\noindent where $w_j$ is the $j$-th word in the word sequence $\bm{w}$, $w_{<j}$ represents the words in front of $w_j$, $p(w_j|w_{<j},\bm{v})$ is the probability of correctly reconstructing $j$-th word according to the previous words $w_{<j}$ and the ground truth word sequence $\bm{w}$, and $\gamma$ and $\delta$ are balance weights.

\noindent\textbf{Hierarchical constraints. }$\rm Y^2Seq2Seq$ employs three constraints on $\bm{F}_s$ and $\bm{F}_t$ to further increase the discriminability of the learned cross-modal representations by employing more detailed discriminative information. These constraints provide hierarchically discriminative information at the class level, the instance pair level, and the instance level.

For the instance pair $(s,t)$ from shape class $c$, the first constraint makes each instance in the pair be correctly classified using a softmax classifier. Here, we use only one softmax classifier to simultaneously classify $\bm{F}_s$ and $\bm{F}_t$ in both modalities. This helps push the two modalities together in the cross-modal representation space. The estimated shape class $c'$ should be the ground truth shape class $c$. Thus, the classification loss is the negative log likelihood as below,

\begin{equation}
L_{\rm C1}=-\log p(c'=c|\bm{F}_s)-\log p(c'=c|\bm{F}_t),
\end{equation}

\noindent where $p(c'=c|\bm{F}_s)$ and $p(c'=c|\bm{F}_t)$ are probabilities of correctly classifying 3D shape $s$ and text $t$, respectively.

We use a triplet constraint~\cite{cvprSchroffKP15} to leverage the relationships between instance pairs, such as $(s^+,t^+)$ and $(s^-,t^-)$. The instance pair level constraint ensures the feature of $s^+$ is closer to the feature of $t^+$ than the feature of any other $s^-$ or $t^-$. Thus, our triplet loss is defined as below,

%\begin{equation}
%\begin{aligned}
%& L_{\rm C}^2=\sum_{k\in\mathcal{K}} \{[\|\bm{F}_{s_k^+}-\bm{F}_{t_k^+}\|_2^2+\|\bm{F}_{s_k^+}-\bm{F}_{t_k^-}\|_2^2+\eta]_+ \\
%& +[\|\bm{F}_{t_k^+}-\bm{F}_{s_k^+}\|_2^2+\|\bm{F}_{t_k^+}-\bm{F}_{s_k^-}\|_2^2+\eta]_+\}, \\
%\end{aligned}
%\end{equation}

\begin{equation}
\begin{aligned}
& L_{\rm C2}=[\|\bm{F}_{s^+}-\bm{F}_{t^+}\|_2^2+\|\bm{F}_{s^+}-\bm{F}_{t^-}\|_2^2+\mu]_+ \\
& +[\|\bm{F}_{t^+}-\bm{F}_{s^+}\|_2^2+\|\bm{F}_{t^+}-\bm{F}_{s^-}\|_2^2+\mu]_+, \\
\end{aligned}
\end{equation}

\noindent where $[\bullet]_+$ is a $max(0,\bullet)$ function, $(s^+,t^+,t^-)$ and $(t^+,s^+,s^-)$ are two triplets in the training set, $\mu$ is a margin that is enforced between instances within pairs and instances across pairs. The two triplets are formed by finding a $(s^-,t^-)$ in another shape class for each $(s^+,t^+)$.

Finally, we include a joint embedding constraint for discriminative information at the instance level. Although instance level information has been involved in the triplet constraint, two instances $s$ and $t$ in the same pair $(s,t)$ are merely required to be as close as possible there. Our joint embedding constraint further toughly pushes the two features $\bm{F}_s$ and $\bm{F}_t$ of the instances $s$ and $t$ to the same point in the cross-modal representation space. Thus, the joint embedding loss is defined as below,

\begin{equation}
L_{\rm C3}=\|\bm{F}_s-\bm{F}_t\|_2^2.
\end{equation}

Finally, the loss $L_{\rm C}$ defined for our hierarchical constraints is shown below, where $\phi$, $\varphi$ and $\psi$ are balance weights for the hierarchical constraints at different levels.

\begin{equation}
L_{\rm C}=\phi L_{\rm C1}+\varphi L_{\rm C2}+\psi L_{\rm C3},
\end{equation}

%\noindent where $\phi$, $\varphi$ and $\psi$ are balance weights for the hierarchical constraints at different levels.

\noindent\textbf{Objective function. }$\rm Y^2Seq2Seq$ is trained to minimize all the above losses. Thus, the objective function is defined as,

\begin{equation}
\begin{aligned}
\min L_{\rm S}+L_{\rm T}+L_{\rm C}.
\end{aligned}
\end{equation}

We learn our learning targets, the cross-modal representations of 3D shape and text, $\bm{F}_s$ and $\bm{F}_t$, by iteratively updating them using back propagated gradients with a learning rate $\varepsilon$, i.e., $\bm{F}_s\gets \bm{F}_s-\varepsilon\times \partial L / \partial \bm{F}_s$ and $\bm{F}_t\gets \bm{F}_t-\varepsilon\times \partial L / \partial \bm{F}_t$.

%\begin{equation}
%\begin{aligned}
%& \bm{F}_s\gets \bm{F}_s+\varepsilon\times \partial L / \partial \bm{F}_s,\\ & %\bm{F}_t\gets \bm{F}_t+\varepsilon\times \partial L / \partial \bm{F}_t.\\
%\end{aligned}
%\end{equation}

\section{Experiments and analysis}
\noindent\textbf{Dataset. }We evaluate $\rm Y^2Seq2Seq$ under the 3D-Text cross-modal dataset by~\cite{chenkevin2018}, which consists of a primitive subset and a ShapeNet subset. Specifically, the primitive subset contains 7,560 shapes and 191,850 descriptions. The ShapeNet subset contains 15,038 shapes and 75,344 descriptions. We employ the same training/test splitting under both subsets  as~\cite{chenkevin2018}.

%In addition, we also evaluate $\rm Y^2Seq2Seq$ under GIF-Text cross-modal dataset~\cite{LiSCTGJL16}. GIF-Text dataset includes 125,781 GIF-Text instance pairs.

%\begin{table}
%  \caption{The balance weights employed in all the three datasets.}
%  \label{table:balanceweights}
%  \centering
%  \begin{tabular}{c|c|c}%llllllllll
%    \hline
%    Weights & Primitive & ShapeNet \\
%    \hline
%    $\alpha(L_{V2V})$ &1&1\\
%    $\beta(L_{V2W})$&2&1\\
%    \hline
%    $\gamma(L_{W2W})$ &0.001&0.01\\
%    $\delta(L_{W2V})$&1&1\\
%    \hline
%    $\phi(L_{\rm C}^1)$&0.0001&0.001\\
%    $\varphi(L_{\rm C}^2)$&0.1($\eta=1$)&0.01($\eta=1.5$)\\
%    $\psi(L_{\rm C}^3)$&0.1&0.1\\
%    \hline		
%  \end{tabular}
%\end{table}

\begin{table}
  \caption{The balance weights employed in all the three datasets.}
  \label{table:balanceweights}
  \centering
  \begin{tabular}{c|cc|cc|ccc}%llllllllll
    \hline
    Set&$\alpha$&$\beta$&$\gamma$&$\delta$&$\phi$&$\varphi$&$\psi$\\
    \hline
    Prim&1&2&0.001&1&0.0001&0.1($\mu=1$)&0.1\\
    Shap&1&1&0.01&1&0.001&0.01($\mu=1.5$)&0.1\\
    %Weights & Primitive & ShapeNet \\
%    \hline
%     &1&1\\
%    &2&1\\
%    \hline
%     &0.001&0.01\\
%    &1&1\\
%    \hline
%    &0.0001&0.001\\
%    &0.1($\eta=1$)&0.01($\eta=1.5$)\\
%    &0.1&0.1\\
    \hline		
  \end{tabular}
\end{table}

\noindent\textbf{Experimental setup. }We first explore how each element of $\rm Y^2Seq2Seq$ affects its performance under the primitive subset and the ShapeNet subset in 3D-Text cross-modal retrieval. Then, we evaluate the performance of $\rm Y^2Seq2Seq$ by comparing it with the state-of-the-art methods in cross-modal retrieval and 3D shape captioning under the two subsets, respectively. In all the experiments, the dimension of the cross-modal representations $\bm{F}_s$ and $\bm{F}_t$ is 128, the dimension of the word embedding $\bm{e}_j$ is set to 512, and both branches in $\rm Y^2Seq2Seq$ are implemented using GRU cells~\cite{Cho2014On}. All the experiments are conducted with a learning rate $\varepsilon$ of 0.00001.

According to the descriptions involved in each subset, we employ dataset-dependent vocabulary. Under the primitive subset, we extract 77 unique words to form the vocabulary. Under the ShapeNet subset, the employed vocabulary is formed by 3587 unique words.
%Under GIF dataset, 11557 unique words consist of the vocabulary.
All the RNNs for descriptions are dynamic, with varying length according to the length of the descriptions.

\begin{table}
  \caption{Effect of coupled ``Y'' like Seq2Seq under primitive subset.}
  \label{table:balance1}
  \centering
  \begin{tabular}{c|cccc|cc|c}%llllllllll
    \hline
      & Metrics & Rec & Pre & R+P &C-$\rm S$& C-$\rm T$& C-$\rm Y$\\
    \hline		
     \multirow{3}{*}{\rotatebox{90}{$\rm S 2 \rm T$}}&RR@1 &1.47 &1.33&1.33&69.87&73.73& \textbf{80.13}\\
     &RR@5 &1.73 &2.67&4.27&70.40&81.60& \textbf{82.53}\\
     &NDCG@5 &1.44 &1.37&1.05&69.87&67.92& \textbf{80.16}\\
    \hline
      \multirow{3}{*}{\rotatebox{90}{$\rm T 2 \rm S$}}&RR@1 &2.27 &2.06&1.85&46.92&72.57&\textbf{92.45}\\
     &RR@5 & 4.70&7.58&3.34&63.37&87.73& \textbf{95.99}\\
     &NDCG@5 &1.76 &3.01&0.97&41.79&70.21& \textbf{88.52}\\
    \hline
  \end{tabular}
\end{table}

\begin{table}
  \caption{Effect of coupled ``Y'' like Seq2Seq under ShapeNet subset.}
  \label{table:balance3}
  \centering
  \begin{tabular}{c|cccc|cc|c}%llllllllll
    \hline
      &Metrics & Rec & Pre & R+P&C-$\rm S$& C-$\rm T$& C-$\rm Y$\\
    \hline		
     \multirow{3}{*}{\rotatebox{90}{$\rm S 2 \rm T$}}&RR@1 &0.07 &0.07&0.13&1.61 &1.74&\textbf{1.88}\\
     &RR@5 &0.34 &0.34&0.34&6.03 &6.17&\textbf{7.51}\\
     &NDCG@5 &0.07 &0.07&0.08&1.44&1.42& \textbf{1.65}\\
    \hline
      \multirow{3}{*}{\rotatebox{90}{$\rm T 2 \rm S$}}&RR@1 &0.13 &0.11&0.07& 0.42&0.77&\textbf{1.04}\\
     &RR@5 &0.34 &0.32&0.35&1.20&3.26& \textbf{4.25}\\
     &NDCG@5 &0.24 &0.21&0.20&0.82&1.98& \textbf{2.62}\\
    \hline
  \end{tabular}
\end{table}

Under each subset, according to the order of magnitude of each loss in $L$, we set both $\alpha$ in $L_{\rm S}$ and $\delta$ in $L_{\rm T}$ to 1, since the losses of reconstructing the view sequence in $\rm S$ and predicting the view sequence in $\rm T$ are similar. Subsequently, all the other balance weights are set based on them. The balance weights used in both subsets are shown in Table~\ref{table:balanceweights}, where the margin $\mu$ involved in the triplet loss is also presented.

%primitives+classification+triplet:shape2txt    88.53    88.80    88.33;  txt2shape    95.99    97.53    95.52

\begin{table}
  \caption{Effect of hierarchical constraints under primitive subset.}
  \label{table:balance2}
  \centering
  \begin{tabular}{c|ccccc}%llllllllll
    \hline
      &Metrics &No& $+L_{\rm C1}$ & $+L_{\rm C1}+L_{\rm C2}$ & $+L_{\rm C}$\\
    \hline		
     \multirow{3}{*}{\rotatebox{90}{$\rm S 2 \rm T$}}&RR@1 &80.13 &83.07&88.53&\textbf{94.13} \\
     &RR@5 &82.53&85.73&88.80&\textbf{94.13} \\
     &NDCG@5 &80.16 &82.43&88.33& \textbf{94.10}\\
    \hline
      \multirow{3}{*}{\rotatebox{90}{$\rm T 2 \rm S$}}&RR@1 &92.45 &93.20&95.99& \textbf{96.66}\\
     &RR@5 &95.99 &97.50&97.53 &\textbf{97.57}\\
     &NDCG@5 &88.52 &89.36&95.52&\textbf{95.87} \\
    \hline
  \end{tabular}
\end{table}

In each cross-modal retrieval experiment below, we show the results in two directions to comprehensively evaluate the performance of $\rm Y^2Seq2Seq$. One direction uses 3D shapes as queries and aims to retrieve instances from the set of descriptions. We call this direction  shape-to-text, which is abbreviated as ``S2T''. In contrast, the other direction uses descriptions as queries and aims to retrieve instances from the set of 3D shapes. Accordingly, we call this direction text-to-shape and abbreviate it as ``T2S''. In addition, the recall rate (RR@$k$)~\cite{chenkevin2018} and NDCG~\cite{Jarvelin02} are used as the metrics.
%The RR@$k$ considers a retrieval successful if at least one sample in the top $k$ retrievals is of the correct class.
In addition, all the results shown in both the ``T2S'' and ``S2T'' directions in the same column in the following tables are obtained with the same trained $\rm Y^2Seq2Seq$.

In each 3D shape captioning experiment below, METEOR~\cite{W14-3348}, ROUGE~\cite{Lin:2004}, CIDEr~\cite{VedantamZP15}, and BLUE~\cite{Papineni2002BMA} are used as the metrics to evaluate the quality of generated descriptions according to the ground truth descriptions, where these metrics are abbreviated as ``M'', ``R'', ``C'', and ``B-1'', ``B-2'', ``B-3'', ``B-4'', respectively in the following tables.

%Specifically, BLUE-1, BLUE-2, BLUE-3, and BLUE-4 are employed. In addition, these METEOR, ROUGE, CIDEr and BLUE are abbreviated as ``M'', ``R'', ``C'', and ``B-1'', ``B-2'', ``B-3'', ``B-4'', respectively in the following tables.

\noindent\textbf{Effect of the coupled ``Y'' like Seq2Seq. }First, we conduct experiments to show how the coupled ``Y'' like Seq2Seq structures contribute to the cross-modal representation learning by joint reconstruction and prediction of view and word sequences.

As shown in Table~\ref{table:balance1} and Table~\ref{table:balance3}, we conduct six experiments under the two subsets, respectively, where RR@1, RR@5 and NDCG@5 are used as metrics in all the experiments. Under each subset, all the six experiments are conducted without hierarchical constraints. The first two experiments are conducted with only reconstruction within the same modality and only with prediction across the modalities, which are indicated as ``Rec'' and ``Pre'' in the tables. In these two experiments, the two ``Y'' like branches are degenerated into line like branches. The third experiment is conducted with both reconstruction and prediction, but the two branches are not coupled, where the decoders of the two branches are not sharing parameters, as indicated as ``R+P'' in the tables. The fourth and fifth experiments are conducted with coupled branches but without joint reconstruction and prediction, i.e., the decoders of both branches regard either the view sequence or the word sequence as the target, as denoted as ``C-$\rm S$'' or ``C-$\rm T$'', respectively. The last experiment is conducted with coupled ``R+P'', as indicated as ``C-$\rm Y$'' in the tables. Note that we use the same balance weights in Table~\ref{table:balanceweights} in all the four experiments.

\begin{table}
  \caption{Effect of hierarchical constraints under ShapeNet subset.}
  \label{table:balance4}
  \centering
  \begin{tabular}{c|ccccc}%llllllllll
    \hline
      &Metrics &No& $+L_{\rm C1}$ & $+L_{\rm C1}+L_{\rm C2}$ & $+L_{\rm C}$\\
    \hline		
     \multirow{3}{*}{\rotatebox{90}{$\rm S 2 \rm T$}}&RR@1 &1.88 &2.82&3.42& \textbf{6.77}\\
     &RR@5 & 7.51&10.19&10.59& \textbf{19.30}\\
     &NDCG@5 & 1.65&2.40&2.59& \textbf{5.30}\\
    \hline
      \multirow{3}{*}{\rotatebox{90}{$\rm T 2 \rm S$}}&RR@1 &1.04 &1.83&1.92& \textbf{2.93}\\
     &RR@5 &4.25 &6.36&6.89 & \textbf{9.23}\\
     &NDCG@5 &2.62 &4.07&4.40& \textbf{6.05}\\
    \hline
  \end{tabular}
\end{table}

Both the results of ``Rec'' and ``Pre'' in Table~\ref{table:balance1} and Table~\ref{table:balance3} show that merely reconstruction within the same modality or prediction across the modalities in the two branches can not learn satisfactory cross-modal representations. This is because the two branches can not bridge the semantic meaning embedded in different modalities. %learn from enough interactions across the modalities to push the modalities together well, even in the ``Pre'' situation.
The ``R+P'' results in the two tables still suffer from the same issue. Even with increasing the interactions across modalities, joint reconstruction and prediction in each branch can only slightly increase the discriminability of the learned cross-modal representations in some metrics. With the two branches coupled, only using the features from one modality as the mapping target of the decoders achieves much better results than the results without coupling, as shown by ``C-$\rm S$'' or ``C-$\rm T$''. This means the coupled branches are able to help bridge the semantic meaning embedded in different modalities by pushing the modalities together better. To explore how to learn more satisfactory cross-modal representations, we propose to jointly perform reconstruction and prediction by the coupled two branches through sharing the parameters in the decoders of both branches. As shown by the ``C-$\rm Y$'' results, the outperforming of other results demonstrates that the reconstruction, the prediction and the coupling can all contribute to the performance, and both the reconstruction and prediction can only contribute based on the coupling.

\begin{table}
  \caption{Effect of voxel resolution under ShapeNet subset.}
  \label{table:balance5}
  \centering
  \begin{tabular}{c|cccc}%llllllllll
    \hline
      &Metrics & $32^3$ & $64^3$ & $128^3$\\
    \hline		
     \multirow{3}{*}{\rotatebox{90}{$\rm S 2 \rm T$}}&RR@1 &6.77 &7.31& \textbf{7.64}\\
     &RR@5 & 19.30 &19.97& \textbf{20.64}\\
     &NDCG@5 & 5.30 &5.43& \textbf{5.48}\\
    \hline
      \multirow{3}{*}{\rotatebox{90}{$\rm T 2 \rm S$}}&RR@1 & \textbf{2.93} &2.37& 2.70\\
     &RR@5 & 9.23 &8.81& \textbf{9.82}\\
     &NDCG@5 & 6.05 &5.61& \textbf{6.27}\\
    \hline
  \end{tabular}
\end{table}

\noindent\textbf{Effect of the hierarchical constraints. }Next, we explore how each hierarchical constraint contributes to the cross-modal representation learning under the two subsets.

As shown in Table~\ref{table:balance2} and Table~\ref{table:balance4}, we conduct three experiments under the primitive subset and ShapeNet subset, respectively, where RR@1, RR@5 and NDCG@5 are used as metrics in all the experiments. Under each subset, each hierarchical constraint is incrementally added starting from ``Coupled ``Y'''' in Table~\ref{table:balance1} or Table~\ref{table:balance3}, which are indicated as ``No'' in the three experiments. These experiments are conducted with only the classification constraint,  indicated as ``$+L_{\rm C1}$'', with the classification and triplet constraints, indicated as ``$+L_{\rm C1}+L_{\rm C2}$'', and with all hierarchical constraints, indicated as ``$+L_{\rm C}$'', respectively. Note that we use the balance weights in Table~\ref{table:balanceweights} in all the three experiments.

\begin{table}
  \caption{The comparison in Cross-modal retrieval under primitive subset.}
  \label{table:primitiveBLEU}
  \centering
  \begin{tabular}{c|cccc}%llllllllll
    \hline
    &Methods&RR@1&RR@5&NDCG@5\\
    \hline
    \multirow{10}{*}{$\rm S 2 \rm T$}%&Random &2.80 &6.53 &1.84 \\
    &ML &24.67  &29.87 &24.38 \\
    &DS & 80.50  &85.87 &80.36 \\
    &MiViSE&17.87&24.13&16.44\\
    &SLR&1.20&2.80&1.15\\
    &LBAT &5.20  &6.13 &5.25 \\
    &LBAM &89.20  &90.53 &89.48 \\
    &FTST &92.00  &92.40 &91.98 \\
    &FMM &93.47  &93.47 &93.47 \\
    &Our &\textbf{94.13} &\textbf{94.13} &\textbf{94.10} \\
    \hline
    \multirow{10}{*}{$\rm T 2 \rm S$}%&Random &0.24 &0.76 &0.27 \\
    &ML &25.93 &57.24 &25.00 \\
    &DS & 81.77  &90.70 &81.29 \\
    &MiViSE&8.21&15.42&6.84\\
    &SLR&4.08&9.49&2.31\\
    &LBAT &5.06  &15.29 &5.92 \\
    &LBAM &91.13  &98.27 &91.90 \\
    &FTST &94.24  &97.55 &95.20 \\
    &FMM &95.07  &\textbf{99.08} &95.51 \\
    &Our &\textbf{96.66} &97.57 &\textbf{95.87} \\
    \hline
  \end{tabular}
\end{table}

%Under both subsets, all the metrics are increasing along with adding an additional hierarchical constraint. However, we also observe some decreased results when using more constraints, such as RR@5 and NDCG@5 in ``T2S''. This is because the primitive includes some shapes with the exact same description, it is hard to meet all the constraints well when the results have been very high. Under the ShapeNet subset, we observe that all the metrics are increasing along with adding an additional hierarchical constraint.

Under both subsets, all the metrics are increasing along with adding an additional hierarchical constraint. The increasing results in terms of all metrics show that each hierarchical constraint is able to contribute to the performance of $\rm Y^2Seq2Seq$. However, the degree of contribution provided by each constraint is slightly different, depending on the dataset. For example, according to the degree of increase, the three constraints contribute almost equally under the primitive subset, while the joint embedding constraint contributes more than the others under the ShapeNet subset.

\noindent \textbf{Effect of voxel resolution. }In the experiments above, the involved 3D shapes are represented by view sequences rendered from voxel representations of the 3D shapes with a resolution of $32^3$. However, $\rm Y^2Seq2Seq$ can learn from higher-resolution voxel representations without increasing the memory requirements of the neural networks because it performs view-based deep learning. This resolves the memory issue of methods that perform learning directly from the voxel representations. We explore the effect of voxel resolution on the performance of $\rm Y^2Seq2Seq$ in cross-modal retrieval under ShapeNet subset. We use views captured from 3D shapes represented by voxel grids with a resolution of $32^3$, $64^3$ and $128^3$.

The grids with higher resolution can provide more geometry detail on the shape surface, which results in more discriminative 3D shape features. This is verified by the continuously increasing results of ``S2T'' in Table~\ref{table:balance5}. However, because of the limited information embedded in the description, voxels with higher resolution could not keep contributing to more discriminative text features, as shown by the results of ``T2S'', although we have got the best ``T2S'' results with the highest resolution of $128^3$.

%Experiments:
%A. Ablation studies:
%1. Same-modal reconstruction
%2. Different-modal prediction
%3. Y^2
%8. Y^2 without coupled
%4. Y^2+classification loss
%5. Y^2+triplet loss
%6. Y^2+embedding loss
%7. Y^2+classification loss+triplet loss+embedding loss

\noindent \textbf{Cross-modal retrieval. }To evaluate the performance of $\rm Y^2Seq2Seq$, we compare our method with the state-of-the-art methods under the primitive subset and the ShapeNet subset in cross-modal retrieval. To conduct a fair comparison, our results are obtained from voxel grids of $32^3$ resolution under the primitive subset and the ShapeNet subset, which is the same as in the other methods.

\begin{table}
  \caption{The comparison in Cross-modal retrieval under ShapeNet subset.}
  \label{table:shapenetBLEU}
  \centering
  \begin{tabular}{c|cccc}%llllllllll
    \hline
    &Methods&RR@1&RR@5&NDCG@5\\
    \hline
    \multirow{10}{*}{$\rm S 2 \rm T$}%&Random &0.07  &0.34 &0.06\\
    &ML &0.13  &0.47 &0.11 \\
    &DS & 0.13  &0.60 &0.13 \\
    &MiViSE&0.20&0.40&0.10\\
    &SLR&0.27&0.40&0.11\\
    &LBAT &0.20  &0.80 &0.12 \\
    &LBAM &0.07  &0.34 &0.07 \\
    &FTST &0.94  &3.69 &0.85 \\
    &FMM &0.83  &3.37 &0.73 \\
    &Our &\textbf{6.77} &\textbf{19.30} &\textbf{5.30} \\
    \hline
    \multirow{10}{*}{$\rm T 2 \rm S$}%&Random &0.11  &0.35 &0.23 \\
    &ML &0.13  &0.61 &0.36 \\
    &DS & 0.12  &0.65 &0.38 \\
    &MiViSE&0.11&0.31&0.20\\
    &SLR&0.11&0.38&0.24\\
    &LBAT &0.04  &0.20 &0.12 \\
    &LBAM &0.08  &0.34 &0.21 \\
    &FTST &0.22  &1.63 &0.87 \\
    &FMM &0.40  &2.37 &1.35 \\
    &Our &\textbf{2.93} &\textbf{9.23} &\textbf{6.05} \\
    \hline
  \end{tabular}
\end{table}

Under the two subsets, we compare our method with metric learning (ML)~\cite{SongXJS16}, deep symmetric structured joint embedding (DS)~\cite{ReedALS16}, MiViSE~\cite{YaleSongGIF2018}, SLR~\cite{xushen2018aaai}, association learning with only TST round trips (LBAT)~\cite{HausserMC17}, and several methods from~\cite{chenkevin2018}, such as LBA-MM (LBAM), Full-TST (FTST), and Full-MM (FMM). Among all these methods, MiViSE, SLR and our method are view-based, and the rest are voxel-based.

\begin{table}
  \caption{The comparison in 3D shape captioning under primitive subset.}
  \label{table:BLEUp}
  \centering
  \begin{tabular}{c|c|c|c|c|c|c|c}%llllllllll
    \hline
    Model&M&R&C&B-1&B-2&B-3&B-4\\
    \hline
    %Rand&0.244&0.546&0.213&0.520&0.377&0.273&0.204\\
    %Rand&0.24&0.55&0.21&0.52&0.38&0.27&0.20\\
    %SLR&&&&&\\
    %SLR-N&0.184&0.439&0.128&0.417&0.309&0.213&0.152\\
    SLR-N&0.18&0.44&0.13&0.42&0.31&0.21&0.15\\
    %MiV-N&0.345&0.686&0.532&0.658&0.534&0.447&0.387\\
    MiV-N&0.35&0.67&0.53&0.66&0.53&0.45&0.39\\
    %S2VT&0.472&0.874&0.956&0.879&0.818&0.754&0.700\\
    S2VT&0.47&0.87&0.96&0.88&0.82&0.75&0.70\\
    \hline
    %Our-N&0.697&0.982&1.371&0.980&0.970&0.962&0.955\\
    Our-N&\textbf{0.70}&\textbf{0.98}&\textbf{1.37}&\textbf{0.98}&\textbf{0.97}&\textbf{0.96}&\textbf{0.96}\\
    Our &0.54&0.92&1.21&0.92&0.88&0.84&0.80\\
    %Our&0.537&0.921&1.212&0.920&0.884&0.844&0.803\\
    \hline
  \end{tabular}
\end{table}

As shown in Table~\ref{table:primitiveBLEU} and Table~\ref{table:shapenetBLEU}, $\rm Y^2Seq2Seq$ outperforms all the other methods in terms of all metrics. Particularly, our results significantly outperform other methods under the ShapeNet subset. For example, our result is about 6 times better in ``S2T'' retrieval and about 4 times better in ``T2S'' retrieval than the state-of-the-art FMM. We believe these experimental improvements come from our novel way of bridging the semantic meaning embedded in the cross-modal sequences in the process of joint reconstruction and prediction by the two coupled branches.

%In addition, under GIF dataset, we also compare $\rm Y^2Seq2Seq$ with DCCA~\cite{pmlr-v28-andrew13}, CorrAE~\cite{FangxiangACMM2014}, DeViSE~\cite{NIPS2013_5204}, and MiViSE~\cite{YaleSongGIF2018}. Note that there is no provided shape class information to use, so we set the weight of classification constraint to zero, as shown in Table~\ref{table:balanceweights}.

\begin{figure*}[!htb]
  \centering
  % the following command controls the width of the embedded PS file
  % (relative to the width of the current column)
  %\includegraphics[width=.95\linewidth, bb=39 696 126 756]{figures/definition3.eps}
   \includegraphics[width=\linewidth]{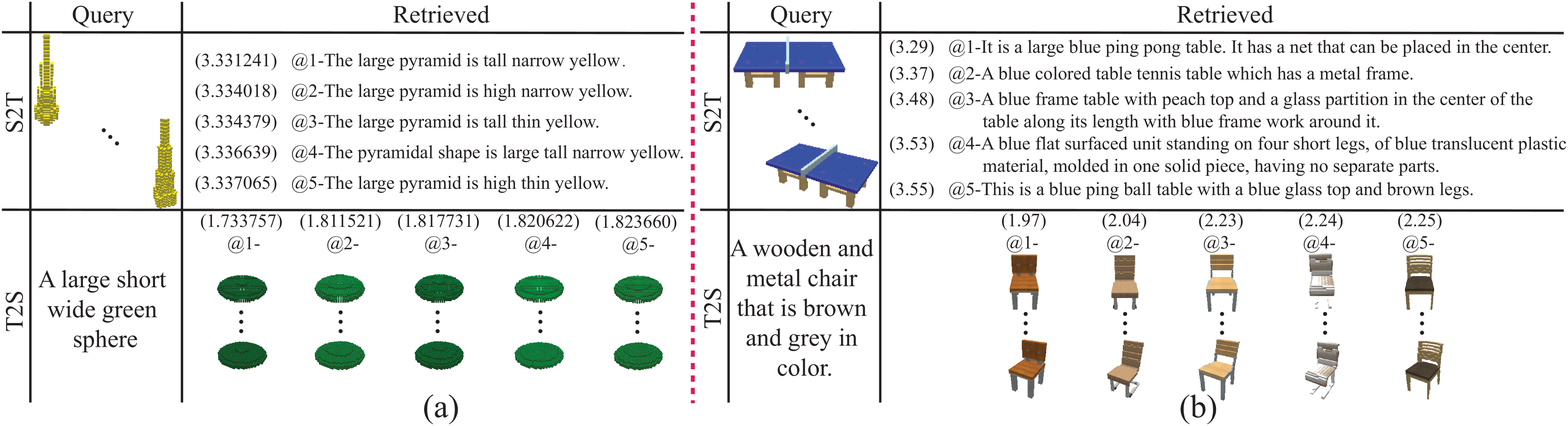}
  % replacing the above command with the one below will explicitly set
  % the bounding box of the PS figure to the rectangle (xl,yl),(xh,yh).
  % It will also prevent LaTeX from reading the PS file to determine
  % the bounding box (i.e., it will speed up the compilation process)
  % \includegraphics[width=.95\linewidth, bb=39 696 126 756]{sampleFig}
  %
  %
\caption{\label{fig:Retrieval_Result} The demonstration of ``S2T'' and ``T2S'' retrieval results under (a) primitive subset and (b) ShapeNet subset. The top 5 retrieved results with their distances to the query are shown.}
\end{figure*}

%\begin{table}
%  \caption{The comparison in Cross-modal retrieval under TGIF dataset.}
%  \label{table:BLEU}
%  \centering
%  \begin{tabular}{c|ccccc}%llllllllll
%    \hline
%    &Methods&RR@1&RR@5&RR@10&NDCG@5\\
%    \hline
%    \multirow{5}{*}{$\rm T 2 \rm S$}&DCCA &0.02 &0.08 & 0.17 & -\\
%    &CorAE &0.05 &0.26 & 0.53 & -\\
%    &DeViSE &0.16 &1.04 & 1.72 & -\\
%    &MiViSE &0.58 &2.00 & 3.74 & -\\
%    &Our & & & & \\
%    \hline
%    \multirow{2}{*}{$\rm S 2 \rm T$}&Ours & & & & \\
%    &Others & & & & \\
%    \hline
%  \end{tabular}
%\end{table}

Furthermore, we visualize the ``S2T'' and the ``T2S'' retrieval results under the primitive subset and the ShapeNet subset in Fig.~\ref{fig:Retrieval_Result} (a) and Fig.~\ref{fig:Retrieval_Result} (b) respectively, where the Top-5 retrieved items with their distances to the query are shown. The retrieved 3D shapes or descriptions highly match their description queries or 3D shape queries. The retrieved results show that our learned discriminative representations can jointly represent 3D shapes and descriptions in the same cross-modal space. In addition, they are able to distinguish the subtle difference between shapes and descriptions. For example, the retrieved shapes in ``T2S'' retrieval under ShapeNet subset are all matching the description query, but there are still some style and appearance differences among them. Similarly, the retrieved descriptions in ``S2T'' retrieval under the primitive subset are all matching the 3D shape query, but the emphasized characteristics of the query 3D shape are still different.

\begin{figure}[!htb]
  \centering
  % the following command controls the width of the embedded PS file
  % (relative to the width of the current column)
  %\includegraphics[width=.95\linewidth, bb=39 696 126 756]{figures/definition3.eps}
   \includegraphics[width=\linewidth]{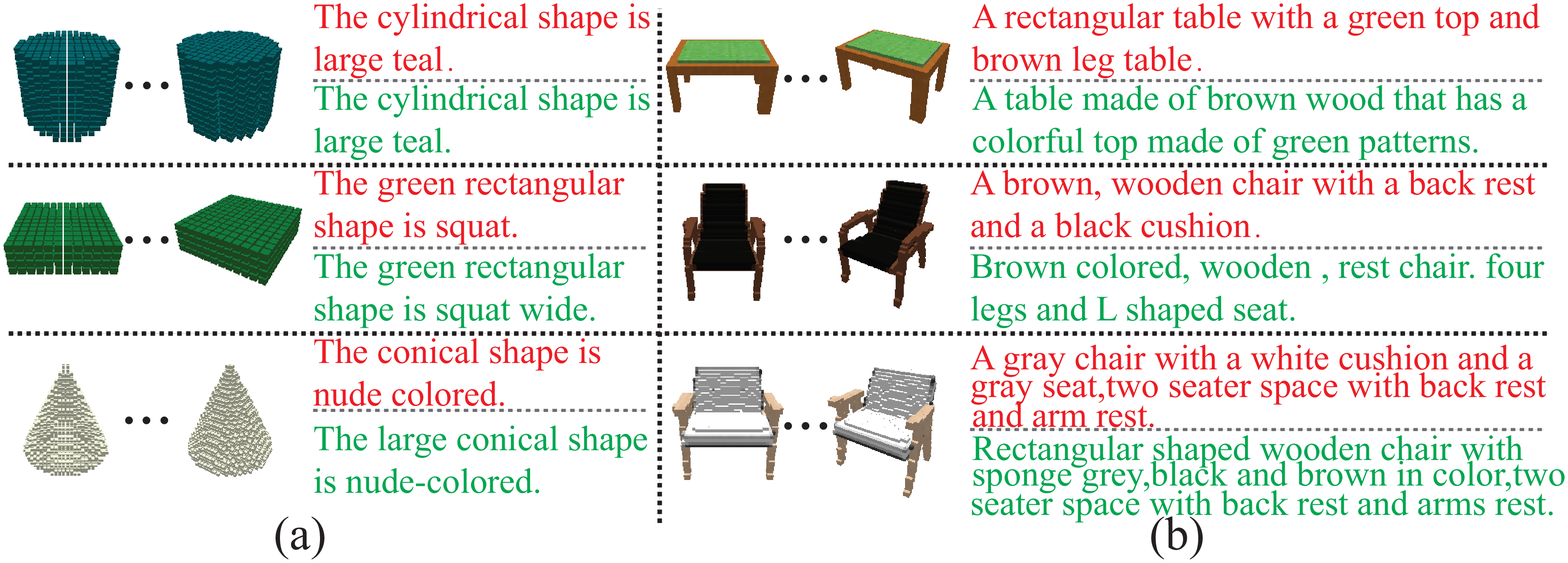}
  % replacing the above command with the one below will explicitly set
  % the bounding box of the PS figure to the rectangle (xl,yl),(xh,yh).
  % It will also prevent LaTeX from reading the PS file to determine
  % the bounding box (i.e., it will speed up the compilation process)
  % \includegraphics[width=.95\linewidth, bb=39 696 126 756]{sampleFig}
  %
  %
\caption{\label{fig:GeneratedDescriptions} The demonstration of generated descriptions under (a) primitive subset, (b) ShapeNet subset, where the ground truth descriptions are shown in green (second one of each pair of descriptions).}
\end{figure}

\noindent \textbf{3D shape captioning. }To further evaluate the performance of $\rm Y^2Seq2Seq$, we compare our method with the state-of-the-art methods under the primitive subset and the ShapeNet subset in 3D shape captioning. For the methods which can directly generate descriptions for 3D shapes, we will directly evaluate the generated descriptions according to the ground truth descriptions. Otherwise, we will use the feature of a 3D shape to retrieve the nearest description of the 3D shape as the generated description, where we use the suffix ``-N'' to denote these results. Note that the captioning results of different methods with either retrieval or generation are produced by the same trained parameter in cross-modal retrieval experiments.

%\begin{table}
%  \caption{The comparison in 3D shape captioning under primitive shape subset.}
%  \label{table:BLEU}
%  \centering
%  \begin{tabular}{c|c|c|c|c|c}%llllllllll
%    \hline
%    Methods&M&B-1&B-2&B-3&B-4\\
%    \hline
%    Random&&&&&\\
%    %SLR&&&&&\\
%    SLR-NN&&&&&\\
%    MiViSE-NN&&&&&\\
%    S2VT&&&&&\\
%    FMM-NN&&&&&\\
%    \hline
%    Our-NN &&&&&\\
%    Our&&&&&\\
%    \hline
%  \end{tabular}
%\end{table}

\begin{table}
  \caption{The comparison in 3D shape captioning under ShapeNet subset.}
  \label{table:BLEUs}
  \centering
  \begin{tabular}{c|c|c|c|c|c|c|c}%llllllllll
    \hline
    Model&M&R&C&B-1&B-2&B-3&B-4\\
    \hline
    %Rand&0.115&0.260&0.044&0.401&0.170&0.070&0.027\\
    %Rand&0.12&0.26&0.04&0.40&0.17&0.07&0.03\\
    %SLR&&&&&\\
    %SLR-N&0.112&0.244&0.046&0.396&0.172&0.078&0.038\\
    SLR-N&0.11&0.24&0.05&0.40&0.17&0.08&0.04\\
    %MiV-N&0.159&0.363&0.140&0.613&0.350&0.208&0.123\\
    MiV-N&0.16&0.36&0.14&0.61&0.35&0.21&0.12\\
    %S2VT&0.212&0.453&0.274&0.673&0.428&0.256&0.150\\
    S2VT&0.21&0.45&0.27&0.67&0.43&0.26&0.15\\
    \hline
    Our1-N&0.22&0.41&0.29&0.57&0.34&0.22&0.17\\
    %Our-N&0.216&0.408&0.292&0.574&0.341&0.224&0.168\\
    Our1 &\textbf{0.29}&\textbf{0.56}&\textbf{0.71}&\textbf{0.80}&\textbf{0.65}&\textbf{0.53}&\textbf{0.46}\\
    %Our&0.293&0.557&0.713&0.798&0.647&0.531&0.455\\
    \hline
    Our2-N&0.22&0.41&0.30&0.57&0.34&0.23&0.18\\
    %Our-N&0.218&0.408&0.297&0.567&0.342&0.231&0.178\\
    Our2 &\textbf{0.30}&\textbf{0.56}&\textbf{0.72}&\textbf{0.80}&\textbf{0.65}&\textbf{0.54}&\textbf{0.46}\\
    %Our&0.297&0.563&0.724&0.801&0.653&0.537&0.460\\
    \hline
    Our3-N&0.22&0.41&0.31&0.58&0.35&0.24&0.19\\
    %Our-N&0.223&0.414&0.313&0.576&0.351&0.239&0.185\\
    Our3 &\textbf{0.29}&\textbf{0.55}&\textbf{0.70}&\textbf{0.80}&\textbf{0.64}&\textbf{0.52}&\textbf{0.44}\\
    %Our&0.294&0.552&0.704&0.793&0.635&0.516&0.438\\
    \hline
  \end{tabular}
\end{table}

Under the primitive subset and the ShapeNet subset, we compare our method with SLR~\cite{xushen2018aaai}, MiV~\cite{YaleSongGIF2018}, and S2VT~\cite{Venugopalan2015}. We show our results obtained by voxel grids with different resolutions, such as $32^3$, $64^3$ and $128^3$ under the ShapeNet subset, where these three sets of results are denoted as ``Our1'', ``Our2'' and ``Our3'' respectively in Table~\ref{table:BLEUs}. We use the description predictor $P^W$ in the 3D shape branch $\rm S$ to generate a description of a 3D shape. In addition, we also show our results obtained by nearest neighbor retrieval.

From the results shown, we can see that all our results obtained by either retrieval or generation under the primitive subset are the best. In addition, our results with different voxel resolutions obtained by either retrieval or generation under the ShapeNet subset are also the best among all competitors.

In addition, although all our results obtained by generation are good, they are always no better than the ones by retrieval under the primitive subset, while this phenomenon is reversed under the ShapeNet subset, as shown by the results with different voxel resolutions. This is because the primitive subset only contains simple primitive shapes that are described by simple sentences with similar patterns. $\rm Y^2Seq2Seq$ obtains good cross-modal retrieval results, as shown in Table~
\ref{table:primitiveBLEU}, which makes the retrieved descriptions almost the same as the ground truth. In contrast, the ShapeNet subset contains complex shapes with complex descriptions. It is not easy to retrieve the right description. However, $\rm Y^2Seq2Seq$ can still generate some words for the key characteristics of the shape, such as color and material, which leads to better results with generation.

Moreover, we find the voxel resolution only slightly increases the results with both retrieval and generation. This phenomenon is similar to the one in the ``T2S'' retrieval in Table~\ref{table:balance5}, where higher resolution mainly contributes to ``S2T'' retrieval but only provides a small contribution to ``T2S'' retrieval.

%Under GIF subset, we compare our method with DCC~\cite{HendricksVRMSD16}, SAT~\cite{ChenLCHFS17}, S2VT~\cite{Venugopalan2015}. In addition, we also show our results obtained by nearest searching.

Finally, we also visualize the description generated by $\rm Y^2Seq2Seq$ under the primitive subset and the ShapeNet subset in Fig.~\ref{fig:GeneratedDescriptions}. Compared to the ground truth, the 3D shape branch $\rm S$ of $\rm Y^2Seq2Seq$ can generate high quality description for the characteristics embedded in the view sequence according to the knowledge learned from joint reconstruction and prediction in the two branches. $\rm Y^2Seq2Seq$ generates almost the same descriptions as the ground truth under the primitive subset, while it generates descriptions with similar semantic meanings to the ground truth under the ShapeNet subset. The generated descriptions successfully represent the class, shape, color, and materials of the 3D shapes, which indicates that $\rm Y^2Seq2Seq$ can jointly understand 3D shapes and text quite well.

%\begin{table}
%  \caption{The comparison in 3D shape captioning under GIF dataset.}
%  \label{table:BLEU}
%  \centering
%  \begin{tabular}{c|c|c|c|c|c}%llllllllll
%    \hline
%    Methods&M&B-1&B-2&B-3&B-4\\
%    \hline
%    DCC& 11.8 &34.6& 17.5&9.3&4.1\\
%    SAT& 14.5 &47.5 & 29.2 &17.9&10.3\\
%    S2VT& 16.1 & 51.1& 31.3&19.1&11.2\\
%    \hline
%    Ours-NN&&&&&\\
%    Ours& & & & &\\
%    \hline
%  \end{tabular}
%\end{table}

\section{Conclusions}
To resolve the issue of cubic complexity of 3D voxel grids for 3D-Text cross-modal representation learning, we propose $\rm Y^2Seq2Seq$ as a view-based model to learn cross-modal representations by joint reconstruction and prediction of view and word sequences. The novel coupled ``Y''-like Seq2Seq structures effectively bridge the semantic meaning embedded in the 3D shape and text modalities by simultaneously reconstructing within the same modality and predicting across modalities. In addition, more detailed discriminative information can be successfully employed to further increase the discriminability of learned cross-modal representations by our novel hierarchical constraints. Our experimental results show that $\rm Y^2Seq2Seq$ is able to learn more discriminative cross-modal representations from high-resolution voxels for 3D-Text cross-modal retrieval and 3D shape captioning compared to the competing state-of-the-art techniques.

\section{Acknowledgments}
Yu-Shen Liu is the corresponding author. This work was supported by National Key R\&D Program of China (2018YFB0505400), the National Natural Science Foundation of China (61472202), and Swiss National Science Foundation grant (169151). We thank all anonymous reviewers for their constructive comments.

%B. Parameter comparison:
%Variation of weights of each loss
%
%C. Comparison with other methods on simple dataset (SRL)
%D. Comparison with other methods on complex dataset
%E. Comparison with other methods on GIF-TEXT dataset
%F. Visualization of learned embedding

\bibliographystyle{aaai}
\bibliography{../paper}

\end{document}